\documentclass[11pt,a4paper]{article}
\usepackage[hyperref]{eacl2021}
\usepackage{times}
\usepackage{latexsym}

\usepackage{graphicx}
\usepackage{array}
\usepackage{mathtools}
\usepackage{subcaption}
\usepackage{pbox}
\usepackage{makecell}
\usepackage{float}
\usepackage[group-separator={,},group-minimum-digits={3}]{siunitx}
\newcolumntype{P}[1]{>{\centering\arraybackslash}p{#1}}

\usepackage{microtype}

\aclfinalcopy 
\def\aclpaperid{***} 

\title{Exploring Early Prediction of Buyer-Seller Negotiation Outcomes}

\author{Kushal Chawla$^1$, Gale Lucas$^1$, Jonathan May$^2$, Jonathan Gratch$^1$ \\
  University of Southern California \\
  Los Angeles, USA \\
  $^1$\texttt{chawla,lucas,gratch@ict.usc.edu} \\
  $^2$\texttt{jonmay@isi.edu}
 }

\date{}

\begin{document}
\maketitle
\begin{abstract}
Agents that negotiate with humans find broad applications in pedagogy and conversational AI. Most efforts in human-agent negotiations rely on restrictive menu-driven interfaces for communication. To advance the research in language-based negotiation systems, we explore a novel task of early prediction of buyer-seller negotiation outcomes, by varying the fraction of utterances that the model can access. We explore the feasibility of early prediction by using traditional feature-based methods, as well as by incorporating the non-linguistic task context into a pretrained language model using sentence templates. We further quantify the extent to which linguistic features help in making better predictions apart from the task-specific price information. Finally, probing the pretrained model helps us to identify specific features, such as trust and agreement, that contribute to the prediction performance.
\end{abstract}

\section{Introduction}

Research shows that students graduating from high school, universities, and even MBA programs are under-prepared in key interpersonal competencies~\cite{rubin2009relevant}. The usual way to teach negotiation skills is through in-class simulations, which is expensive. Automated systems can dramatically reduce the costs of, and increase access to, negotiation training. Such agents can also help to advance the negotiation capabilities of conversational assistants~\cite{leviathan2018google}. However, most efforts in human-machine negotiations rely on restrictive communication interfaces such as a pre-defined menu of options~\cite{mell2016iago}. Such interfaces are unrealistic and introduce a cognitive load for humans, limiting their usability. Instead, language-based interactions are desirable as they allow negotiators to express themselves freely, better capturing realistic aspects like diversity and rapport building.
\begin{table}[t!]
\scalebox{0.6}{
\begin{tabular}{ll}
\hline
\multicolumn{2}{l}{\textbf{Single speed bianchi practically new (Bike)}} \\
\multicolumn{2}{l}{\textbf{Listing: $\$300$, Buyer Target: $\$150$, Agreed (Ground-truth): $\$200$}} \\ \hline
\textbf{Negotiation Seen by the Model} & \textbf{Predictions} \\ \hline
\pbox{9.1cm}{\vspace{0.1cm}\textbf{Buyer:} Hi. I am interested in your bicycle. How long have you had it for?\\\textbf{Seller:} I have had it for a little over a month.\vspace{0.1cm}} & $\$225.91$ ($f$=$0.2$) \\ \hline
\pbox{9.1cm}{\vspace{0.1cm}\textbf{Buyer:} Is there anything wrong with it?\\\textbf{Seller:} Nothing wrong at all pretty much new.\vspace{0.1cm}} & $\$227.37$ ($f$=$0.4$) \\ \hline
\pbox{9.1cm}{\vspace{0.1cm}\textbf{Buyer:} Okay. I see that you are listing it at $\$300$. However, I can buy a new one for that. Honestly, without any sort of warranty available and the fact that it is used-I can do $\$150$.\\\textbf{Seller:} It actually still has over $10$ months of the warranty that comes with the bike when you buy it. I will not go as low as $\$150$ I can do $\$225$ though.\vspace{0.1cm}}& $\$196.70$ ($f$=$0.6$) \\ \hline
\pbox{9.1cm}{\vspace{0.1cm}\textbf{Buyer:} Usually a warranty doesn't transfer if you sell it. I can do $\$200$.\\\textbf{Seller:} If you have any problems with it, within the next $10$ months save my number. I can do $\$200$ you will pick up tonight?\vspace{0.1cm}} & $\$195.58$ ($f$=$0.8$) \\ \hline
\pbox{9.1cm}{\vspace{0.1cm}\textbf{Buyer:} Sure, I can do that.\\\textbf{Seller:} (OFFER $\$200$)\\\textbf{Buyer:} (ACCEPT)\vspace{0.1cm}} & $\$201.71$ ($f$=$1.0$) \\ \hline
\end{tabular}}
\caption{\label{tab:example} Sample negotiation along with BERT-based model predictions of the final agreed price obtained by looking at the first $f$ fraction of the negotiation.}
\vspace{-0.6cm}
\end{table}
However, this makes the setting more complex and it's unclear how an agent can use the linguistic clues from utterances while engaging in a negotiation. To this end, we explore the prediction of final negotiation outcomes with access to only a partial dialogue. Specifically, we investigate buyer-seller negotiations~\cite{he2018decoupling} by varying the fraction of utterances that a model can access: $f$$\in$$\{0.2, 0.4, 0.6, 0.8, 1.0\}$. Such early prediction mechanisms allow an agent to evaluate the consequent impact of the dialogue seen till now, which can guide its negotiation strategy. Moreover, predicting the final outcomes, as the negotiation unfolds, can enable an agent to provide critical feedback to their opponent, which has been shown to improve social skills in individuals~\cite{monahan2018autonomous}. Along with investigating the feasibility of the early prediction problem, we study whether language features improve the model performance, along with other task-specific meta information.

We now list our contributions. 1) We define a novel problem of predicting negotiation outcomes from a fraction of the dialogue (Section \ref{sec:setup}). Prior work has extensively investigated negotiations~\cite{adair2001negotiation, koit2018people}. Tackling an easier problem, \citet{nouri2013prediction} predict intermediate outcomes using hand-crafted multi-modal features. However, to the best of our knowledge, we are the first to computationally study language cues in the early phases of negotiation that direct the final outcomes. 2) As traditional methods provide better interpretability, we design a variety of models from heuristics to feature-based to neural for this task. We define a metric called \texttt{Accuracy} $\pm$ $k$, measuring the \%age of cases where the model performs ‘well’, as controlled by $k$. This allows us to compare our methods in a useful manner (Section \ref{sec:expts}). 3) We show how to integrate BERT~\cite{devlin2019bert} with the non-linguistic task-context to adapt it for complex negotiation data (Section \ref{sec:approach}). The model performs at par with the best performing non-neural ensemble method, trained on hand-crafted features developed over many years. 4) We design ablation studies and a probing technique for BERT to pinpoint which linguistic features help more than others for early outcome prediction task. We show a sample negotiation along with model predictions in Table \ref{tab:example}.

\section{Problem Setup}
\label{sec:setup}
We study buyer-seller negotiations, a key research area in the literature~\cite{williams2012iamhaggler}. First, we describe our problem setup and key terminologies by discussing the dataset used, and later, we formalize our problem definition.

\noindent\textbf{Dataset:} We use the Craigslist Bargaining dataset (\textbf{CB}) introduced by \citet{he2018decoupling}. The dataset consists of $6682$ dialogues ($9.2$ turns on average) between buyers and sellers who converse to negotiate the price of a given product. The products come from $1402$ ad postings belonging to six categories: phones, bikes, housing, furniture, car, and electronics. Each posting contains the title, category, a target for the seller (called listing price), and a secret target for the buyer (called target price). The final price after the agreement is called the agreed price, which we aim to predict. 

\noindent\textbf{Defining the problem:} Given a product scenario $S$: (\textit{Category}, \textit{Title}, \textit{Listing Price}, \textit{Target Price})\footnote{Assume buyer's perspective where target price is known.}, define the interactions between a buyer and seller using an ordered sequence of $N$ events $E_N:$\textless$e_{1}, e_{2}, ..., e_{N}$\textgreater. Event $e_{i}$ is a tuple: (\textit{Sender}, \textit{Type}, \textit{Data}). \textit{Sender} is either buyer or seller, \textit{Type} can be one of (message, offer, accept, reject or quit) and \textit{Data} is either a text message, offer price or can be empty depending on the \textit{Type}. Nearly $80\%$ of events in \textbf{CB} dataset are of type `message'. An offer is usually made and accepted at the end of each negotiation. Since the offers directly contain the agreed price (which we want to predict), we only consider `message' events in our models. Given the scenario $S$ and the first $f$ fraction of the events $E_{fN}$, our problem is then to learn the function $F_{f}$: $A = F_{f}(S, E_{fN})$ where $A$ refers to the final agreed price between the two negotiating parties.
\section{Methodology}
\label{sec:approach}
We start by presenting the design of feature-based approaches. Later, we show how we adapt BERT for our task.
\subsection{Hand-crafted feature-based methods}
\noindent\textbf{Task-specific features (TSF):} These features primarily come from the meta information associated with each negotiation. We consider $9$ \textbf{TSFs} comprising (a) $6$-dimensional one-hot encoding of the product category, (b) buyer target price, and (c) the most recent price quoted by both negotiators in the utterances that the model can access.

\noindent\textbf{Language Features (LF):} We leverage various affect lexicons from prior work to extract linguistic features from raw utterances (Table \ref{tab:featSets}). Further, we extract several syntactic features based on aggregate word-level statistics, POS and NER tags, and text readability measures from the Python NLTK toolkit. This results in nearly $150$ features that are extracted separately from buyer's and seller's utterances. Table \ref{tab:featSets} presents the feature sets, counts, and the most significant features based on an F-regression test. Consistent with our intuition based on prior work studying the role of affect in negotiations~\cite{barry2004laughed, olekalns2009mutually}, many features such as Warriner-Valence, LIWC-Power, and PERMA-Relationship are significant at $p$\textless$0.05$. We provide the complete list of features in the Appendix.

\noindent\textbf{Modeling:} Combining all the features into a single vector performs the best for us. For modeling, we consider several traditional models, namely, Linear Regression, Support Vector Regression, Feed-Forward, Decision Tree and Random Forest.
\begin{table}[t!]
\centering
\scalebox{0.6}{
\begin{tabular}{p{5.3cm}p{0.5cm}P{5.5cm}}
\hline
\textbf{Feature Set} & \textbf{Count} & \textbf{Top $\mathbf{3}$ predictive features} \\ \hline
\textbf{Syntactic} & $88$ & Is-a-question(B)$^*$,
Is-difficult-word(S)$^*$, $\#$words(B)$^*$\\
\textbf{Formality}~\cite{brooke2010automatic} & $4$ & Informal(S)$^*$, Informal(B)$^*$, Formal(S)$^*$  \\
\textbf{Temporal}~\cite{park2015living} & $6$ & Is-present-word(B)$^*$, Is-past-word(B), Is-present-word(S)\\
\textbf{Warriner}~\cite{warriner2013norms} & $18$ & Arousal-low(B)$^*$, Valence-high(B)$^*$, Arousal-high(S)$^*$  \\
\textbf{PERMA}~\cite{seligman2012flourish} & $20$ & Pos-P(B)$^*$, Pos-R(B)$^*$, Pos-A(B)$^*$ \\
\textbf{EmoLex}~\cite{mohammad2013nrc} & $20$ & Surprise(B)$^*$, Positive(S)$^*$, Anticipation(B)$^*$ \\
\textbf{LIWC}~\cite{pennebaker2001linguistic} & $146$ & Sad(S)$^*$, Reward(B)$^*$, Power(S)$^*$  \\ \hline
\end{tabular}}
\caption{\label{tab:featSets} Description of the extracted linguistic features. Predictive power is based on F-regression test where $^*$ denotes significance with $p$\textless$0.05$. B or S denotes extraction from either Buyer's or Seller's text.}
\end{table}
\begin{table}[tb]
\centering
\scalebox{0.65}{
\begin{tabular}{ll}
\hline
\textbf{Available Information} & \textbf{Sentence Template} \\ \hline
\multicolumn{2}{c}{\textbf{Scenario}} \\ \hline 
$<$Category$>$ &  Category is $<$Category$>$.\\
$<$Target$>$ & Target Price is $<$Target$>$.\\
$<$Title$>$ & Title is $<$Title$>$.\\ \hline
\multicolumn{2}{c}{\textbf{Events (Sender, Type, Data)}} \\ \hline
(Buyer, message, $<$message$>$) & Buyer: $<$message$>$\\
(Seller, message, $<$message$>$) & Seller: $<$message$>$\\ \hline
\end{tabular}}
\caption{\label{tab:info2nl} Natural language templates to represent the auxiliary information for effective consumption by the BERT encoder.}
\end{table}
\subsection{Adapting BERT for negotiations}
Pre-trained language models like BERT~\cite{devlin2019bert} have achieved strong performance on a wide range of NLP tasks. However, since our setup deals with various auxiliary pieces (category, price, etc.), we cannot directly leverage these models since they have been trained only on natural language inputs. Moreover, training BERT further on such structured data before using it is challenging given the lack of enough in-domain data. Hence, we propose using simple predefined sentence templates to incorporate the auxiliary information into the same embedding space. Table \ref{tab:info2nl} shows how we represent the category, target price, and the title in natural language sentences which are concatenated to form our scenario $S$. We also define templates to capture the negotiator identity and the conveyed message. The scenario $S$ and the dialogue messages are all concatenated together using a separator token to form the final BERT input. We alternate between a sequence of $0$s and $1$s for segment embeddings to differentiate between the scenario and each individual turn. We found such inclusions to be crucial for model performance. [CLS] token embedding is a contextualized encoding for the entire token sequence. We pass this encoding through a two-layer feed-forward network to finally predict the agreed price. The model is fine-tuned in an end-to-end manner using the mean squared error between the predicted price and the ground-truth.
\section{Experiments}
\label{sec:expts}
We primarily answer three questions: (a) Is it feasible to predict final outcomes without observing the complete dialogue? (b) To what extent does the natural language incorporation help in the prediction? (c) What language features contribute to this prediction performance? We compare a variety of methods by training them with different proportions of the negotiation seen, namely, $f \in \{0.2, 0.4, 0.6, 0.8, 1.0\}$.

\noindent\textbf{Training Details:} We handle the price variance by normalizing all prices by \textit{Listing Price}. The output is unnormalized before evaluation. We remove dialogues that do not reach agreement or are noisy. We further explain these steps in Appendix, which result in $3854$ training instances, $451$ for validation, and $630$ for test. We use BERT-base~\cite{devlin2019bert} due to multiple segments in our model input and small data size. The training was done for $10$k iterations with Adam and a batch size of $4$. We performed extensive hyperparameter tuning by varying the hidden units, activation, dropout, learning rate, and weight decay by a combination of grid, randomized, and manual search. We describe tuning and validation statistics in the Appendix.

\noindent\textbf{Evaluation:} We define \texttt{Accuracy}$\pm$ $k$ as the \%age of cases where the prediction lies within $k$\% of the target. We use $k$=$10$ for our experiments. We find \texttt{Accuracy}$\pm$ $10$ more relevant, interpretable, and less susceptible to outliers than other regression metrics like MAE. Hence, we use \texttt{Accuracy}$\pm$ for all evaluations. Regardless, we include results on MAE in the Appendix for completeness.

\noindent\textbf{Results:} Table \ref{tab:testacc2} summarizes our results. \textbf{AAP} ignores the input and directly outputs the \textbf{A}verage \textbf{A}greed \textbf{P}rice over the training data. Low performance of \textbf{AAP} attests to the challenging problem of high variance across diverse product categories. \textbf{AAP-n} is the normalized version with average taken after normalizing the \textit{Agreed Price} by \textit{Listing Price} and then unnormalizing at inference time. It achieves a respectable $56.7\%$ \texttt{Accuracy}$\pm10$, justifying our price normalization criteria. It even beats both \textbf{Listing} and \textbf{Target Price}, where we just output the associated \textit{Listing} or \textit{Target Price}. Higher performance of \textbf{Target Price} compared to \textbf{Listing Price} suggests that buyers controlled the negotiation to their advantage in \textbf{CB} dataset. Among feature-based methods, \textbf{Random Forest (RF)} performs the best as depicted in Table \ref{tab:traditionalmodels}. Using language features (\textbf{LF)} over just the task-specific features (\textbf{TSF}) results in better performance for all five fractions, indicating the impact of linguistic cues on negotiation outcomes. Remarkably, the model achieves $75.3\%$ just after looking at $0.6$ fraction of the dialogue. We implement a Bidirectional-LSTM and train it from scratch with the same input format as BERT for comparison. However, the model fails to perform on this task. \textbf{BERT-FT} adaptation which fine-tunes the BERT model, highly improves over \textbf{Bi-LSTM}, attesting to the need for incorporating pre-trained representations. For early prediction, \textbf{RF} performs even better than \textbf{BERT-FT}. We attribute this high performance of \textbf{RF} to 1) \textbf{RF} is an ensemble model of $100$ decision trees, and 2) local decision making in trees which helps in tackling outliers in our dataset. Regardless, we observe that this gap is almost nullified by \textbf{BERT-Ensemble}, which uses a $1$-layer MLP to combine the predictions of all BERT models with $f$ lesser than and equal to the current one.\footnote{For eg., ensemble at $0.4$ combines the outputs from the BERT models trained for $f$ equal to $0.2$ and $0.4$.}. More sophisticated methods of incorporating the non-linguistic context or other pre-trained models such as RoBERTa~\cite{liu2019roberta} and XLNet~\cite{yang2019xlnet} may further boost the performance. We encourage future work to adapt them for negotiations in the future.
\begin{table}[t!]
\centering
\scalebox{0.7}{
\begin{tabular}{lllllll}    
\hline
\textbf{Model} & \multicolumn{5}{c}{\textbf{Fraction of the dialogue seen \textbf{(}$f$\textbf{)}}} & \textbf{Mean} \\
& $\mathbf{0.2}$ & $\mathbf{0.4}$ & $\mathbf{0.6}$ & $\mathbf{0.8}$ & $\mathbf{1.0}$ &\\ \hline
\textbf{AAP} & $3.9$ & $3.9$ & $3.9$ & $3.9$ & $3.9$ & $3.9$\\ 
\textbf{AAP-n} & $\mathbf{56.7}$ & $\mathbf{56.7}$ & $\mathbf{56.7}$ & $\mathbf{56.7}$ & $\mathbf{56.7}$ & $\mathbf{56.7}$\\ 
\textbf{Listing Price} & $20.9$ & $20.9$ & $20.9$ & $20.9$ & $20.9$ & $20.9$\\
\textbf{Target Price} & $49.1$ & $49.1$ & $49.1$ & $49.1$ & $49.1$ & $49.1$\\ \hline
\textbf{RF: TSF} & $65.8$ & $66.3$ & $74.8$ & $86.0$ & $87.6$ & $76.1$\\
\textbf{RF: TSF+LF} & $\mathbf{67.2}$ & $\mathbf{69.1}$ & $\mathbf{75.3}$ & $\mathbf{88.8}$ & $\mathbf{89.5}$ & $\mathbf{78.0}$ \\ \hline
\textbf{Bi-LSTM} & $58.9$ & $55.1$ & $54.0$ & $59.0$ & $56.2$ & $56.6$\\
\textbf{BERT-FT} & $\mathbf{66.6}$ & $67.9$ & $73.9$ & $91.4$ & $92.4$ & $78.4$\\ 
\textbf{BERT-Ensemble} & $\mathbf{66.6}$ & $\mathbf{68.9}$ & $\mathbf{74.3}$ & $\mathbf{91.7}$ & $\mathbf{92.5}$ & $\mathbf{78.8}$\\ \hline
\end{tabular}}
\caption{\label{tab:testacc2} Performance on \texttt{Accuracy}$\pm10$ on the negotiation outcome prediction task. RF: Random Forest, AAP: Average Agreed Price.}
\end{table}
\begin{table}[t!]
\centering
\scalebox{0.7}{
\begin{tabular}{lllllll}
\hline
\textbf{Model} & \multicolumn{5}{c}{\textbf{Fraction of the dialogue seen \textbf{(}$f$\textbf{)}}} & \textbf{Mean}\\
& $\mathbf{0.2}$ & $\mathbf{0.4}$ & $\mathbf{0.6}$ & $\mathbf{0.8}$ & $\mathbf{1.0}$ & \\ \hline
\textbf{Linear Regression} & $63.0$ & $64.4$ & $66.1$ & $67.3$ & $67.3$ & $65.6$\\
\textbf{SVR(kernel$=$'rbf')} & $60.6$ & $62.4$ & $60.4$ & $63.1$ & $63.3$ & $62.0$\\
\textbf{Feed-Forward} & $52.4$ & $50.9$ & $49.9$ & $50.7$ & $50.6$ & $50.9$\\
\textbf{Decision Tree} & $51.8$ & $54.2$ & $61.4$ & $82.2$ & $80.2$ & $66.0$\\ 
\textbf{Random Forest} & $\mathbf{67.2}$ & $\mathbf{69.1}$ & $\mathbf{75.3}$ & $\mathbf{88.8}$ & $\mathbf{89.5}$ & $\mathbf{78.0}$ \\ \hline
\end{tabular}}
\caption{\label{tab:traditionalmodels} Performance for different traditional models on TSF+LF features.}
\end{table}

\noindent\textbf{Ablation Analysis:} Table \ref{tab:langfeatureAblation} presents the ablation analysis for \textbf{RF}. The drop in performance suggests a higher contribution of \textbf{Syntactic} and \textbf{LIWC} over \textbf{Lex} features (from the remaining lexicons).
\begin{table}[t!]
\centering
\scalebox{0.7}{
\begin{tabular}{lllllll}
\hline
\textbf{Model} & \multicolumn{5}{c}{\textbf{Fraction of the dialogue seen \textbf{(}$f$\textbf{)}}} & \textbf{Mean} \\
& $\mathbf{0.2}$ & $\mathbf{0.4}$ & $\mathbf{0.6}$ & $\mathbf{0.8}$ & $\mathbf{1.0}$ &\\ \hline
\textbf{RF: TSF+LF} & $\mathbf{67.2}$ & $\mathbf{69.1}$ & $\mathbf{75.3}$ & $\mathbf{88.8}$ & $\mathbf{89.5}$ & $\mathbf{78.0}$\\ \hline
$-$\textbf{Syntactic} & $64.7$ & $67.6$ & $74.2$ & $87.6$ & $89.4$ & $76.7$\\
$-$\textbf{LIWC} & $66.3$ & $67.6$ & $75.0$ & $87.5$ & $89.2$ & $77.1$\\
$-$\textbf{Lex} & $65.2$ & $68.7$ & $74.5$ & $88.3$ & $\mathbf{89.5}$ & $77.2$\\ \hline
\end{tabular}}
\caption{\label{tab:langfeatureAblation} Random Forest (RF) ablation performance on \texttt{Accuracy}$\pm10$ for Syntactic, LIWC, and Lex features (from rest of the lexicons).}
\end{table}
\begin{table}[t!]
\centering
\scalebox{0.65}{
\begin{tabular}{lll}
\hline
\textbf{$f$} & \textbf{Buyer} & \textbf{Seller} \\ \hline
$\mathbf{0.2}$ &Home, Neg-A, Space&Trust, $\#$Adv, Arousal-low \\
$\mathbf{0.4}$ &Has-if, Is-a-question, Num-words & Agreement, Is-future, Pronouns  \\
$\mathbf{0.6}$ & Is-a-question, I, Dominance-high & Is-stress-word, I, We \\ \hline
\end{tabular}}
\caption{\label{tab:bertProbingNumbers} Top $3$ language features captured by BERT, as depicted by the probing analysis. Larger table with MAE scores in the Appendix.}
\vspace{-0.3cm}
\end{table}

\noindent\textbf{Probing BERT:} Probing has been useful in analyzing the BERT pre-trained model~\cite{tenney2019you, liu2019linguistic,rogers2020primer}. To extend the analysis for our task, we design a probing technique: predict language features (\textbf{LF}) from the [CLS] representation gathered before and after training for our task and observe the improvements. The predictor is a naive $1$-layer (size=$100$) feed-forward network, ensuring that the performance is largely a consequence of contextualized BERT representations. We take the positive improvement in performance as evidence that a feature is being captured by BERT for our task. Table \ref{tab:bertProbingNumbers} presents the top features which see an improvement in MAE in early prediction cases. The presence of trust, agreement, and various syntactic properties is consistent with our intuition from prior work~\cite{olekalns2009mutually} and ablation results, indicating that the model is capturing useful information. 

\section{Conclusion and Future Work}
\label{sec:conclusion}
We study early outcome prediction in buyer-seller price negotiations. We find that it is feasible to reasonably predict negotiation outcomes with limited access to the dialogue utterances. Further, language features help to make better predictions of the final agreed price even in the early stages of a negotiation and these features can be identified using the devised probing technique. In the future, we plan to use the early prediction model as a plugin to guide the development of negotiation dialogue systems.

\bibliography{eacl2021}
\bibliographystyle{acl_natbib}

\appendix

\section{Data Pre-processing}
Before doing any modeling or feature extraction, we perform a few pre-processing steps on the publicly available Craigslist Bargaining (CB) dataset. These must be performed for reproducing the results.

\begin{enumerate}
    \item \textbf{Removing inconclusive negotiations:} We only consider the negotiations where an agreement was reached. These are the instances for which the ground truth is available ($\sim75\%$ of the data). 
    \item \textbf{Removal of outliers:} To mitigate the effect of outliers, we only use the training instances with a normalized agreed price within $0.3$ and $2.0$. Most of the dialogues beyond this range were manually verified to be erroneous. For instance, in some cases, the buyer and seller see different products in front of them, which leads to a nonsensical dialogue between them or in some cases, the reported agreed price does not reflect the actual price negotiated between the two participants. 
    \item \textbf{Noise reduction:} To reduce noise, we remove all dialogues where the subjects talked about the meta-data of the task rather than taking part in the negotiation. This search was done through a set of bad keywords: (`hit', `negotiat', `requester', `turk') and search results were manually verified to be noisy.
    \item \textbf{Price Normalization:} As mentioned in the main paper, all the price information in the input and output is normalized by the corresponding listing price. This is done to tackle the huge variance in the price values, especially across products in different categories. For the input side, this normalization is done on target price and every price quoted in the negotiation dialogue. \textbf{For proper consumption by BERT's sub-word level tokenizer}, three steps are performed for each price value, specifically on the input side: (a) The price is normalized by listing price, (b) The resulting number is rounded to $3$ digits, and (c) The number is multiplied by $1000$. These steps help to easily capture numbers in BERT for our case. In order to validate whether BERT is able to map the tokenized price information to the actual regression output, we built a dummy input containing only the actual agreed price, which we want to predict. This input was again normalized using the above steps. We observed that BERT was able to map the dummy input to the output perfectly, achieving \texttt{Accuracy}$\pm10$ in the high nineties. This observation motivated us to capture the auxiliary information in the same space and not rely on any additional price or feature embeddings.
\end{enumerate}
After pre-processing, we end up with $3854$ training dialogues, while $451$ and $630$ for validation and testing respectively.
\section{Hand-crafted Features}
We provide a complete list of features used for our analysis in Table \ref{tab:featsetscomplete}. \textbf{Task specific features (TSFs)} comprise the product and price information in the dialogue. Each listed language feature resulted in two feature values: one extracted from the buyer's text in the negotiation and one from the seller. These features are together referred to as \textbf{Language Features} or \textbf{LF}.

Language features (\textbf{LF}) can be classified into two sub-types: (a) Word-level counts, and (b) aggregate sentence-level scores. Most of the features belong to the first sub-type. However, text quality metrics and VAD scores belong to type (b). While most of the features are based on lexicons collected from an extensive review of the negotiation and affective computing literature, we build many syntactic features manually based on a review of close to $50$ samples in the training data.

For all the count-based features, we perform stemming on each word in the dialogue. We combine these stems with the original words in the dialogue before feature computation. We scale all the features to have mean $0$ and a standard deviation of $1$. 

We perform the F-regression test using the Scikit-learn toolkit. This is used only to assess the correlations of the features with the final outcomes and not to filter the feature set. We achieve the best performance on the whole set of $302$ features.
\section{Training details}
In this section, we provide several key training details, hoping they would promote the reproducibility of our results.

\noindent\textbf{Computing Infrastructure:} All neural network experiments were performed on a single Nvidia GeForce RTX $2070$ GPU with $8$ GB of RAM. The non-neural approaches were directly trained on the CPU.

\noindent\textbf{Implementation Framework:} Neural methods were implemented in PyTorch using the HuggingFace Transformers toolkit for the implementation of pre-trained BERT-base model and the corresponding tokenizer. The non-neural techniques were based on the Scikit Learn package in Python. 

\noindent\textbf{Bi-LSTM model implementation:} The Bi-LSTM model uses the same input as BERT. The input was tokenized using the same tokenizer as BERT from the HuggingFace toolkit. However, the input is first passed through a randomly initialized embedding matrix, which is then passed to a bidirectional LSTM network. Similar to BERT's implementation, the output of bi-LSTM then passes through a two-layer feed-forward network, finally predicting the agreed price. The model is end to end trained by using the Mean Squared Error (MSE) Loss. We discuss the hyperparameters involved and provide tuning details below.
    
\noindent\textbf{Hyper-parameter tuning:} Table \ref{tab:trainingDetails} lists the hyper-parameter bounds, the number of search trials, search technique, search metric, and the approximate runtime for the implemented methods. The hyper-parameters not mentioned in the table were used in their default settings, depending on the implementation framework.

\noindent\textbf{Best-performing Hyperparameters:} Table \ref{tab:best-params} reports the best performing hyperparameters and the corresponding number of parameters for our neural models trained with different fractions of the data.

\noindent\textbf{Validation performance statistics:} Table \ref{tab:validationResults} lists the validation performance on \texttt{Accuracy}$\pm10$ corresponding to the best performing hyperparameters. Wherever available, we also report the mean and the standard deviation of the validation performance. For the BERT-based method, the hyperparameter tuning was a combination of randomized and manual search. Given the longer runtime, several hyperparameter combinations were manually tried for a smaller number of iterations, while randomized search was done for a few other cases. The mean and standard deviations, in this case, are not clearly defined, and hence, have not been reported. 

\section{Test performance on MAE}
While our primary evaluation metric is \texttt{Accuracy}$\pm10$, we report results on MAE in Table \ref{tab:testmae} for completeness. 
\begin{table}[ht!]
\centering
\scalebox{0.65}{
\begin{tabular}{llllll}
\hline
\textbf{Model} & \multicolumn{5}{c}{\textbf{Fraction of the dialogue seen}} \\ \hline
& $\mathbf{0.2}$ & $\mathbf{0.4}$ & $\mathbf{0.6}$ & $\mathbf{0.8}$ & $\mathbf{1.0}$ \\ \hline
\textbf{AAP} & $1878.42$ & $1878.42$ & $1878.42$ & $1878.42$ & $1878.42$ \\
\textbf{AAP-n} & $172.04$ & $172.04$ & $172.04$ & $172.04$ & $172.04$ \\
\textbf{Listing Price} & $323.81$ & $323.81$ & $323.81$ & $323.81$ & $323.81$ \\
\textbf{Target Price} & $237.37$ & $237.37$ & $237.37$ & $237.37$ & $237.37$ \\ \hline
\textbf{RF: TSF} & $125.73$ & $121.65$ & $104.14$ & $79.83$ & $77.04$ \\
\textbf{RF: TSF+LF} & $132.61$ & $125.13$ & $106.54$ & $73.77$ & $77.57$ \\ \hline
\textbf{Bi-LSTM} & $164.30$ & $176.89$ & $194.04$ & $162.50$ & $176.14$ \\
\textbf{BERT} &$126.81$ & $137.47$ & $110.39$ & $66.48$ & $61.64$ \\
\textbf{BERT-Ensemble} & $126.81$ & $128.72$ & $107.61$ & $66.66$ & $61.28$ \\ \hline
\end{tabular}}
\caption{\label{tab:testmae} MAE performance on test set.}
\end{table}
\section{BERT probing analysis}
The probing method uses a naive $1$-layer feedforward network, having $100$ hidden units. We used Adam solver with a batch size of $16$. The model was trained for $5$ epochs. We used early stopping with a tolerance of $1e^{-3}$. The rest of the parameters were used in the default setting as per the Scikit Learn implementation.

To provide further clarity, for every language feature mentioned in Table \ref{tab:featsetscomplete}, the probing architecture was trained two times: one using the corresponding [CLS] representation from the pre-trained BERT model as the input, and another using the same [CLS] representation after the model was trained on our negotiation outcome prediction task. Before using these representations, we applied normalization to bring every input dimension to mean $0$ and a standard deviation of $1$.

Table \ref{tab:probingnumbers} presents the MAE scores for the top features sorted based on the percentage improvement in MAE after training BERT. Top $3$ features for buyer and seller from this Table are presented in the main paper.

\begin{table*}[ht!]
\centering
\scalebox{0.82}{
\begin{tabular}{p{6cm}p{11cm}}
\hline
\textbf{Feature Set} & \textbf{List of features} \\ \hline
\multicolumn{2}{c}{\textbf{Task Specific Features (TSF)}} \\
\textbf{Product Category} & Is-electronics, Is-phone, Is-housing, Is-bike, Is-furniture, Is-car \\
\textbf{Price Features} & Buyer target price, Buyer last price quoted, Seller last-price quoted \\ \hline
\multicolumn{2}{c}{\textbf{Language Features (LF) (For both Buyer and Seller)}} \\
\textbf{Syntactic} & Number of turns, Number of words, Word-length, I, You, We, They, Is-question, Exclaimation, Dot, Caps, Thank, Number of Sentences, Is-alpha, Is-numeric, Is-high-quantifier, Is-low-quantifier, Agreement, Disagreement, Has-if, Is-stress-word, Has-final, Apology words, Number of verb, Number of adj, Number of noun, Number of pron, Number of adv, Number of adp, Number of conj, Number of det, Number of punct, Number of person, Number of org, Number of gpe, flesch-reading-ease, smog-index, flesch-kincaid-grade, coleman-liau-index, automated-readability-index, dale-chall-readability-score, is-difficult, linsear-write-formula, gunning-fog \\ 
\textbf{Warriner} & Valence-low, Valence-high, Arousal-low, Arousal-high, Dominance-low, Dominance-high, Valence-score, Arousal-score, Dominance-score\\
\textbf{Formality} & Formal, Informal \\
\textbf{PERMA} & Neg-P, Neg-E, Neg-R, Neg-M, Neg-A, Pos-P, Pos-E, Pos-R, Pos-M, Pos-A \\
\textbf{Temporal} & Is-present, Is-past, Is-future \\ 
\textbf{EmoLex} & Anger, Anticipation, Disgust, Fear, Joy, Negative, Positive, Sadness, Surprise, Trust\\
\textbf{LIWC} & Achievement, Adjectives, Adverbs, Affect, Affiliation, Anger, Anx, Articles, Assent, Auxiliary Verbs, Biological Processes, Body, Causal, Certainty, Cognitive Processes, Comparisons, Conjunctions, Death, Differentiation, Discrepancies, Drives, Family, Feel, Female, Filler Words, Future Focus, Past Focus, Present Focus, Friends, Function Words, Health, Hear, Home, I, Informal Language, Ingest, Insight, Interrogatives, Impersonal Pronouns, Leisure, Male, Money, Motion, Negations, Negative Emotions, Netspeak, Nonfluencies, Numbers, Perceptual Processes, Positive Emotions, Power, Personal Pronouns, Prepositions, Pronouns, Quantifiers, Relativity, Religion, Reward, Risk, Sad, See, Sexual, She, He, Social, Space, Swear, Tentative, They, Time, Verbs, We, Work, You\\ \hline
\end{tabular}}
\caption{\label{tab:featsetscomplete} Complete list of features used in our analysis. Each language feature was computed twice, one from the buyer's text, one from the seller's text.}
\end{table*}
\begin{table*}[ht!]
\centering
\scalebox{0.75}{
\begin{tabular}{lp{5cm}llll}
\hline
\textbf{Model} & \textbf{Hyperparameter bounds} & \textbf{Search Trials} & \textbf{Search Technique} & \textbf{Search metric} & \textbf{Runtime}\\ \hline
\textbf{Random Forest} & Estimators: ($50$, $100$, $200$), Maximum Depth: ($10$, $50$, $100$, $500$, No limit) & $15$ & Grid & \texttt{Accuracy}$\pm10$ & \textless $1$ minute\\ 
\textbf{Bi-LSTM} & LSTM hidden units: ($50$, $250$, $500$), Learning Rate: ($10^{-3}$, $10^{-4}$, $10^{-5}$), Activation: (relu, tanh, sigmoid, linear), Adam Weight Decay ($0$, $10^{-3}$), Dropout: ($0$, $0.1$)  & $10$ & Randomized & \texttt{Accuracy}$\pm10$ & \textless $10$ minutes \\ 
\textbf{BERT} & Fully-connected layer hidden units: ($50$, $100$, $200$), Learning Rate: ($10^{-3}$, $10^{-4}$, $10^{-5}$), Activation: (relu, tanh, sigmoid, linear), Adam Weight Decay ($0$, $0.001$), Dropout: ($0$, $0.1$) & NA & Randomized + Manual & \texttt{Accuracy}$\pm10$ & \textless $30$ minutes \\ 
\textbf{BERT-Ensemble} & Hidden layer size: ($50$, $100$, $200$), Activation functions: (relu, tanh) & $10$ & Randomized& \texttt{Accuracy}$\pm10$ & \textless $1$ minute\\ \hline
\end{tabular}}
\caption{\label{tab:trainingDetails} Training details for different models discussed in the paper. Note that these hyper-parameters were separately optimized for each fraction of the data. The hyper-parameters not listed here were used in their default settings, depending on the implementation framework.}
\end{table*}
\begin{table*}[ht!]
\centering
\scalebox{0.65}{
\begin{tabular}{p{4cm}|p{6.5cm}|p{2.2cm}|p{6.5cm}|p{2.2cm}}
\hline
\textbf{Fraction of the data seen} & \multicolumn{2}{c|}{\textbf{Bi-LSTM}} & \multicolumn{2}{c}{\textbf{BERT}} \\ \hline & \textbf{Best Hyperparameters} & \textbf{$\#$parameters} & \textbf{Best Hyperparameters} & \textbf{$\#$parameters}\\ \hline
$\mathbf{0.2}$ & Hidden units:$500$, Learning rate:$10^{-3}$, Weight Decay:$0$, Dropout:$0$, Activation:sigmoid, Embedding size:$100$ & $3,326,501$ & Hidden units:$250$, Learning rate:$10^{-5}$, Weight Decay:$0$, Dropout:$0$, Activation:relu& $109,674,741$\\
$\mathbf{0.4}$ &Hidden units:$250$, Learning rate:$10^{-3}$, Weight Decay:$10^{-3}$, Dropout:$0.1$, Activation:tanh, Embedding size:$200$& $2,293,701$ & Hidden units:$50$, Learning rate:$10^{-5}$, Weight Decay:$10^{-3}$, Dropout:$0.1$, Activation:relu& $109,520,741$ \\
$\mathbf{0.6}$ &Hidden units:$500$, Learning rate:$10^{-4}$, Weight Decay:$0$, Dropout:$0.1$, Activation:tanh, Embedding size:$100$& $3,659,501$ & Hidden units:$250$, Learning rate:$10^{-5}$, Weight Decay:$0$, Dropout:$0$, Activation:relu& $109,674,741$ \\
$\mathbf{0.8}$ &Hidden units:$250$, Learning rate:$10^{-3}$, Weight Decay:$0$, Dropout:$0.1$, Activation:sigmoid, Embedding size:$50$& $1,131,201$ & Hidden units:$500$, Learning rate:$10^{-5}$, Weight Decay:$0$, Dropout:$0$, Activation:relu& $109,867,241$ \\
$\mathbf{1.0}$ &Hidden units:$500$, Learning rate:$10^{-3}$, Weight Decay:$0$, Dropout:$0$, Activation:relu, Embedding size:$200$& $4,929,201$ & Hidden units:$50$, Learning rate:$10^{-5}$, Weight Decay:$0$, Dropout:$0.1$, Activation:tanh& $109,520,741$\\ \hline
\end{tabular}}
\caption{\label{tab:best-params} Best performing hyper-parameters and the corresponding number of parameters for our neural methods, corresponding to each fraction of the data used for training. The parameters for non-neural methods can be directed found in the respective Scikit-learn package.}
\end{table*}
\begin{table*}[ht!]
\centering
\scalebox{0.7}{
\begin{tabular}{llllll}
\hline
\textbf{Model} & \multicolumn{5}{c}{\textbf{Fraction of the dialogue seen}} \\
& $\mathbf{0.2}$ & $\mathbf{0.4}$ & $\mathbf{0.6}$ & $\mathbf{0.8}$ & $\mathbf{1.0}$ \\ \hline
\textbf{AAP} & ($1.11$, $1.11$, $0$) & ($1.11$, $1.11$, $0$) & ($1.11$, $1.11$, $0$) & ($1.11$, $1.11$, $0$) & ($1.11$, $1.11$, $0$) \\ 
\textbf{AAP-n} & ($55.21$, $55.21$, $0$) & ($55.21$, $55.21$, $0$) & ($55.21$, $55.21$, $0$) & ($55.21$, $55.21$, $0$) & ($55.21$, $55.21$, $0$) \\ 
\textbf{Listing Price} & ($24.17$, $24.17$, $0$) & ($24.17$, $24.17$, $0$) & ($24.17$, $24.17$, $0$) & ($24.17$, $24.17$, $0$) & ($24.17$, $24.17$, $0$) \\ 
\textbf{Target Price} & ($54.77$, $54.77$, $0$) & ($54.77$, $54.77$, $0$) & ($54.77$, $54.77$, $0$) & ($54.77$, $54.77$, $0$) & ($54.77$, $54.77$, $0$) \\ \hline
\textbf{RF: TSF} & ($63.63$, $62.52$, $0.64$) & ($64.52$, $62.73$, $0.87$) & ($74.05$, $70.89$, $1.32$) & ($85.14$, $84.52$, $0.38$) & ($86.70$, $86.06$, $0.34$)\\ 
\textbf{RF: TSF+LF} & ($63.85$, $62.75$, $0.68$) & ($63.85$, $63.10$, $0.51$) & ($75.17$, $73.52$, $0.81$) & ($86.70$, $86.10$, $0.41$) & ($88.69$, $88.01$, $0.37$) \\ \hline
\textbf{Bi-LSTM} & ($59.64$, $37.42$, $22.47$) & ($56.31$, $41.35$, $19.61$) & ($55.21$, $42.06$, $19.50$) & ($59.86$, $41.15$, $23.24$) & ($57.87$, $31.35$, $24.01$) \\
\textbf{BERT} & ($65.41$, NA, NA)  & ($67.18$, NA, NA) & ($73.83$, NA, NA) & ($86.69$, NA, NA) & ($87.80$, NA, NA) \\
\textbf{BERT-Ensemble} & ($65.41$, $65.41$, $0$)  & ($66.96$, $66.03$, $0.61$) & ($73.83$, $72.92$, $0.72$) & ($87.58$, $86.83$, $0.35$) & ($88.03$, $87.54$, $0.29$) \\ \hline
\end{tabular}}
\caption{\label{tab:validationResults} Validation performance on \texttt{Accuracy}$\pm10$. The values in the brackets refer to the Best validation performance, Mean performance and the Standard Deviation from left to right. `NA' refers to `Not Available'.}
\end{table*}

\begin{table*}[ht!]
\centering
\scalebox{0.7}{
\begin{tabular}{ll|ll|ll}
\hline
\multicolumn{2}{c|}{$\mathbf{0.2}$} & \multicolumn{2}{c|}{$\mathbf{0.4}$} & \multicolumn{2}{c}{$\mathbf{0.6}$} \\ \hline
\textbf{Feature} & \textbf{MAE scores} & \textbf{Feature} & \textbf{MAE scores} & \textbf{Feature} & \textbf{MAE scores}\\ \hline
Home(B) & $0.062$ : $0.061$ : $0.052$ & Agreement(S) & $0.027$ : $0.538$ : $0.026$ & Is-stress-word(S) & $0.006$ : $0.006$ : $0.005$ \\ 
Trust(S) & $0.078$ : $0.086$ : $0.076$ & Has-if(B) & $0.007$ : $0.115$ : $0.006$ & Is-question(B) & $0.035$ : $0.039$ : $0.033$ \\
$\#$adv(S) & $0.04$ : $0.043$ : $0.038$ & Is-question(B) & $0.046$ : $0.556$ : $0.038$ & I(S) & $0.098$ : $0.107$ : $0.092$ \\
Neg-A(B) & $0.036$ : $0.035$ : $0.032$ & Is-future(S) & $0.027$ : $0.353$ : $0.026$ & We(S) & $0.008$ : $0.008$ : $0.007$ \\
Arousal-low(S) & $0.114$ : $0.103$ : $0.095$ & Number of words(B) & $0.553$ : $6.121$ : $0.459$ & Personal Pronouns(S) & $0.124$ : $0.121$ : $0.116$ \\
Impersonal Pronouns(S) & $0.129$ : $0.136$ : $0.126$ & We(B) & $0.012$ : $0.146$ : $0.011$ & I(B) & $0.039$ : $0.036$ : $0.035$ \\
Prepositions(S) & $0.195$ : $0.172$ : $0.16$ & Causal(B) & $0.083$ : $1.006$ : $0.078$ & Dominance-high(B) & $0.117$ : $0.118$ : $0.115$ \\
Auxiliary Verbs(S) & $0.199$ : $0.202$ : $0.188$ & Pos-P(B) & $0.079$ : $0.902$ : $0.072$ & Motion(S) & $0.053$ : $0.053$ : $0.052$ \\
Number of adj(S) & $0.073$ : $0.064$ : $0.06$ & Leisure(B) & $0.039$ : $0.436$ : $0.036$ & Pos-R(B) & $0.059$ : $0.058$ : $0.057$ \\
Disgust(S) & $0.041$ : $0.035$ : $0.033$ & Pronouns(S) & $0.205$ : $2.205$ : $0.194$ & - & -\\ \hline
\end{tabular}}
\caption{\label{tab:probingnumbers} Top features for early prediction cases, as depicted by the BERT probing analysis. (B) or (S) denotes extraction from either Buyer's or Seller's text. MAE scores in the format (X : Y : Z) represent the following: X: MAE for the Average baseline, where the model just predicts the average output in the training dataset, Y: MAE obtained using [CLS] representations before training BERT, Z: MAE obtained after training BERT. The features for a particular fraction are sorted based on the percentage improvement in Z, over Y.}
\end{table*}

\end{document}